\documentclass{article}

\PassOptionsToPackage{numbers, sort&compress}{natbib}
\usepackage[preprint]{neurips_2026}

\usepackage[utf8]{inputenc}
\usepackage[T1]{fontenc}
\usepackage{hyperref}
\usepackage{url}
\usepackage{booktabs}
\usepackage{amsfonts}
\usepackage{fancyvrb}
\usepackage{amsmath}
\usepackage{amssymb}
\usepackage{graphicx}
\usepackage{xcolor}
\usepackage{pifont}
\usepackage{tikz}
\usetikzlibrary{arrows.meta, positioning, calc, backgrounds, fit, shapes.geometric, shapes.misc}
\usepackage{algorithm}
\usepackage{algpseudocode}

\newcommand{\JP}{\mathsf{JP}}
\newcommand{\EI}{\mathsf{EI}}
\newcommand{\EC}{\mathsf{EC}}
\newcommand{\Approve}{\mathsf{approve}}
\newcommand{\Reject}{\mathsf{reject}}
\newcommand{\Escalate}{\mathsf{escalate}}
\newcommand{\Abstain}{\mathsf{abstain}}
\newcommand{\Ready}{\mathsf{Ready}}
\newcommand{\Boundary}{\mathsf{Boundary}}
\newcommand{\Scope}{\mathsf{Scope}}
\newcommand{\Valid}{\mathsf{Valid}}

\title{Verifiable Agentic Infrastructure: Proof-Derived Authorization for Sovereign AI Systems}

\author{
    Jun He \\
    OpenKedge.io \\
    \texttt{junhe@openkedge.io} \\
    \and
    Deying Yu \\
    OpenKedge.io \\
    \texttt{deying@openkedge.io}
}

\begin{document}

\maketitle

\begin{abstract}
Modern cloud and enterprise systems rely on identity-centric authorization, assuming that callers possessing valid credentials are safe to execute commands. The emergence of autonomous AI agents invalidates this assumption: agents can generate syntactically valid but semantically unsafe actions, making standing privileges a significant operational risk. This risk becomes especially acute in sovereign AI systems, where autonomous agents may interact with cloud infrastructure, regulated data, financial workflows, and national-scale digital services. Governed mutation substrates reduce this risk by interposing on agent actions: agents submit intents, infrastructure evaluates context and policy, and execution is mediated. However, this shifts the trust boundary: how can the decision to authorize an intent be made verifiable, distributed, and replayable?

We introduce a Distributed Trust Framework (DTF), a verification framework for governed mutation systems that computes execution authority from structured, verifiable artifacts. DTF introduces a Justification Proof to encode the admissibility basis of an action, a consensus model for independent evaluation, an ephemeral Execution Identity derived from the approved proof, and an append-only Evidence Chain that preserves the authorization lifecycle. Under stated substrate assumptions, this architecture enforces a compact authorization invariant: no high-stakes execution without a proof object, no derived authority without consensus, and no valid mutation detached from evidence.

We define the model, instantiate it over an OpenKedge-based governed mutation substrate, and show how it maps onto cloud-native environments. By shifting authorization from standing identity to proof-derived authority, DTF provides an infrastructure foundation for making agentic execution governable, auditable, and bounded in sovereign AI deployments.
\end{abstract}

\section{Introduction}

Modern cloud and enterprise systems rely on identity and access management models that assume callers are rational and trustworthy. In these systems, authorization is fundamentally identity-centric: if a service account possesses sufficient permissions, its requested actions are admitted. This model breaks down when applied to autonomous AI agents. Agents generate actions non-deterministically; they can produce API calls that are syntactically valid but semantically unsafe. Granting them broad, standing permissions creates operational risk.

OpenKedge addresses part of this risk by reframing agent-driven mutation as an intent-governed process~\cite{openkedge2026}. Instead of mutating state directly, agents submit proposals. The infrastructure evaluates these proposals against context and policy before allowing execution. OpenKedge is the motivating and concrete substrate in this paper, but the verification problem is broader than any single substrate.

Intent governance leaves a second authorization question: what makes the authorization decision itself verifiable, distributed, and replayable? If the system simply intercepts requests, runs a policy check, and issues a token, it leaves a single point of failure and an unauditable decision gap. What is the stable object that evaluators approve? How is temporary authority derived from the approval record rather than from the caller's standing role? What must be persisted so that an auditor can later reconstruct why execution authority existed at all?

We answer these questions with a \emph{Distributed Trust Framework} (DTF) for proof-derived execution. DTF is a verification layer for governed mutation substrates: it assumes that a system already interposes on mutation attempts, and supplies the authorization semantics needed to make approval proof-bound, consensus-gated, and replayable. It adds four verification constructs:
\begin{itemize}
    \item \textbf{Justification Proofs}, structured artifacts that bind intent, context, policy basis, risk, and execution boundary.
    \item \textbf{Consensus validation}, in which independent evaluators attest to the same proof object under explicit governance rules.
    \item \textbf{Execution Identity}, an ephemeral authority token derived from the approved proof rather than from ambient caller privilege.
    \item \textbf{Evidence Chains}, append-only lifecycle records that preserve proof, attestations, authority, execution, and outcome.
\end{itemize}

DTF has a more specific authorization invariant than ordinary access control: a high-stakes mutation is valid only if its authority can be replayed from recorded proof and approval. In conventional identity and access management, authorization is primarily a property of a principal~\cite{sandhu1992role,hu2015guide}. In zero-trust systems, that principal and its context are checked continuously~\cite{rose2020zero}. In DTF, the principal is no longer the main object of trust. The decision lifecycle is.

The paper keeps mechanisms already covered by OpenKedge brief. Section~\ref{sec:openkedge-delta} states the boundary precisely: we instantiate DTF over an OpenKedge-based governed mutation substrate while defining a general authorization-verification model for such substrates.

Our contributions are:
\begin{itemize}
    \item a proof-derived authority model for governed agentic mutation;
    \item a consensus semantics for independent approval of authorization artifacts;
    \item a definition of Execution Identity as computed, ephemeral authority;
    \item safety and audit properties for proof-bound, consensus-gated, evidence-preserved execution;
    \item a practical cloud mapping that uses existing temporary credential and audit primitives.
\end{itemize}

\section{Related Work}

\paragraph{Access control and zero trust.}
Role-based and attribute-based access control decide whether a principal may perform an action~\cite{sandhu1992role,hu2015guide}. Zero-trust architecture improves this model through continuous verification of requester and context~\cite{rose2020zero}. DTF shifts the verification target from the requester to the mutation lifecycle: authority is valid only if it is derived from an approved proof.

\paragraph{Automated Reasoning and Verified Permissions.}
Modern authorization systems increasingly rely on automated reasoning and formal methods to analyze access invariants. For example, recent frameworks like AWS Cedar encode authorization policies into SMT-solvable formulas to check security properties and use mechanized proofs in Lean to establish properties of the policy engine~\cite{bogle2024cedar,rungta2024cedar}. These systems primarily reason about policy semantics and static authorization decisions. DTF addresses a complementary problem: it structures the runtime authorization lifecycle of autonomous agents so that transient authority is derived from a recorded proof object, checked by independent evaluators, and bounded under stated substrate assumptions before mutation occurs.

\paragraph{LLM Agent Safety and Tool-Use Authorization.}
The adoption of Large Language Models (LLMs) as autonomous agents has exposed security vulnerabilities, including indirect prompt injection~\cite{greshake2023more} and the ability of agents to exploit systems when granted unchecked tool access~\cite{fang2024llm}. Recent surveys describe the growing architecture of agent tool-use~\cite{wang2023survey}; governing those tools remains unsettled. Early mitigation strategies focused on prompt-level guardrails or strict "human-in-the-loop" approvals for every action. At scale, continuous human oversight becomes a bottleneck. DTF shifts the safety boundary from the prompt to the execution infrastructure. It provides a structured "human-on-the-loop" alternative where independent evaluators---rather than the proposing agent---derive execution authority from structured proof artifacts.

\paragraph{Distributed trust.}
Consensus and Byzantine fault-tolerant systems study agreement under faulty participants~\cite{lamport1982byzantine,castro1999practical}. We do not run consensus over replicated application state. Instead, we use explicit multi-evaluator agreement to decide whether a proof is sufficient to derive execution authority.

\paragraph{Provenance and accountability.}
Provenance, event sourcing, and accountability systems preserve causal history and support replay or audit~\cite{buneman2001and,weitzner2008information,kleppmann2017designing}. The DTF Evidence Chain---an append-only lifecycle record introduced earlier---specializes this lineage for authorization-bearing mutations by recording not only effects, but the proof object and approval path that produced authority.

\paragraph{OpenKedge and Sovereign Agentic Loops.}
OpenKedge provides the concrete intent-governed mutation substrate used in this paper's instantiation~\cite{openkedge2026}. The Sovereign Agentic Loops architecture separately develops the decoupling of reasoning from direct execution~\cite{he2026sovereign}. DTF is a substrate-agnostic verification layer that formalizes proof construction, consensus-backed approval, and execution identity semantics rather than a replacement for these systems.

\section{Distributed Trust Framework}

In traditional architectures, authorization is largely a static property of a principal: an agent or service account acts under a standing role, and requests are granted if they fall within the permissions of that role. This model breaks down when applied to autonomous AI agents. Because an agent's behavior is non-deterministic and its reasoning is prone to error or manipulation, granting it broad standing authority creates operational risk. 

DTF is a general verification model for intent-governed mutation systems. It assumes a substrate that already interposes on mutation attempts: agents propose intents, the infrastructure binds context and policy, and execution is mediated rather than direct. Any substrate with this interposition point can adopt DTF's authorization semantics. OpenKedge is the concrete governed mutation substrate used for the implementation and examples in this paper~\cite{openkedge2026}.

Moving from direct execution to intent governance raises a subsequent verification problem. If intent governance simply means a central policy engine says ``yes'' or ``no'' and issues an API key, the system still has a single point of failure and an unauditable decision gap. How is the decision to authorize a mutation made verifiable? What is the stable object that evaluators approve? How is authority bounded to the approved intent, and how is the authorization basis reconstructed after the fact?

The Distributed Trust Framework provides the verification layer above such governed mutation substrates. It models authority not as a standing permission, but as a \emph{derived state}. A proposed mutation does not execute because an agent or service account already has broad permission. It executes only if a recorded proof is approved by independent evaluators and converted into a bounded execution identity.

\subsection{Model Formulation}

We define the Distributed Trust Framework using a formal state-transition model. Let $t$ denote logical time. 

\paragraph{Sets and Spaces:}
\begin{itemize}
    \item $\mathcal{I}$: Space of canonical intents.
    \item $\mathcal{C}$: Space of authorization-relevant context snapshots.
    \item $\mathcal{P}$: Space of policy bundles.
    \item $\mathcal{J}$: Space of Justification Proofs.
    \item $\mathcal{A}$: Space of attestations.
    \item $\mathcal{G}$: Space of governance metadata.
    \item $\mathcal{D} = \{\Approve, \Reject, \Escalate\}$: The decision space.
    \item $\mathcal{E}$: Space of Execution Identities.
    \item $\mathcal{X}, \mathcal{O}$: Spaces of mutation attempts and observed outcomes, respectively.
\end{itemize}

\paragraph{State Variables:}
At any time $t$, the system processes an intent $I_t \in \mathcal{I}$ against context $C_t \in \mathcal{C}$ and policy $P_t \in \mathcal{P}$. This produces a Justification Proof $\JP_t \in \mathcal{J}$, a vector of attestations $A_t \in \mathcal{A}^n$ from evaluator set $V_t=\{v_1,\dots,v_n\}$, and a decision $D_t \in \mathcal{D}$ using metadata $\Gamma_t \in \mathcal{G}$. A successful decision materializes an Execution Identity $\EI_t \in \mathcal{E}$, leading to an attempt $X_t \in \mathcal{X}$ and outcome $O_t \in \mathcal{O}$, durably recorded in an Evidence Chain $\EC_t$.

\paragraph{Core Functions:}
\begin{itemize}
    \item \textbf{Proof Construction:} $f: \mathcal{I} \times \mathcal{C} \times \mathcal{P} \rightarrow \mathcal{J}$ generates the proof $\JP_t = f(I_t, C_t, P_t)$.
    \item \textbf{Attestation:} $v_i: \mathcal{J} \rightarrow \mathcal{A}$ generates an independent attestation $a_i^t = v_i(\JP_t)$.
    \item \textbf{Consensus Rule:} $q: \mathcal{A}^n \times \mathcal{G} \rightarrow \mathcal{D}$ computes the decision $D_t = q(A_t, \Gamma_t)$.
    \item \textbf{Authority Derivation:} $h: \mathcal{J} \times \mathcal{A}^n \times \mathcal{G} \rightarrow \mathcal{E}$ materializes the bounded identity $\EI_t = h(\JP_t, A_t, \Gamma_t)$ if $D_t = \Approve$.
\end{itemize}

\subsection{Execution Pipeline}

The authorization pipeline transforms an intent into a governed execution record. Algorithm~\ref{alg:dtf-pipeline} gives the ordered path. The central requirement is causal ordering: proof precedes approval, approval precedes authority, and execution is incomplete until evidence is durably recorded.

\begin{algorithm}[ht]
\caption{DTF Authorization and Execution Pipeline}
\label{alg:dtf-pipeline}
\begin{algorithmic}[1]
\Statex \textbf{Input:} Intent $I_t$, Context $C_t$, Policy $P_t$, Evaluator set $V_t$, Metadata $\Gamma_t$
\Statex \textbf{Output:} Evidence Chain $\EC_t$
\State $\JP_t \gets f(I_t, C_t, P_t)$ \Comment{Construct Justification Proof}
\For{each evaluator $v_i \in V_t$}
    \State $a_i^t \gets v_i(\JP_t)$ \Comment{Generate independent attestations}
\EndFor
\State $A_t \gets (a_1^t, \dots, a_n^t)$
\State $D_t \gets q(A_t, \Gamma_t)$ \Comment{Evaluate consensus rule}
\If{$D_t = \Approve$}
    \State $\EI_t \gets h(\JP_t, A_t, \Gamma_t)$ \Comment{Derive bounded Execution Identity}
    \State $B_t \gets \Boundary(\JP_t)$
    \If{$\Scope(\EI_t) \preceq B_t$}
        \State $X_t, O_t \gets \text{Execute}(\EI_t)$
    \Else
        \State $X_t, O_t \gets \emptyset, \text{Escalated (Boundary Violation)}$
    \EndIf
\Else
    \State $\EI_t \gets \emptyset$
    \State $X_t, O_t \gets \emptyset, \text{Rejected or Escalated}$
\EndIf
\State $\EC_t \gets (I_t, C_t, P_t, \JP_t, A_t, \Gamma_t, D_t, \EI_t, X_t, O_t)$
\State \Return $\EC_t$
\end{algorithmic}
\end{algorithm}

\subsection{Threat Model and Assumptions}

We assume the autonomous agent proposing intents is untrusted. Its internal reasoning, prompts, and local state may be non-deterministic, hallucinated, or compromised by prompt injection. The Trusted Computing Base (TCB) consists of the DTF verification layer (Justification Engine, Evaluation Swarm, Execution Broker, Evidence Store) and the underlying governed mutation substrate (e.g., OpenKedge). We assume the TCB is uncompromised and the substrate correctly interposes on execution. The framework defends against rogue or unsafe agent actions, not against total compromise of the verification infrastructure itself.

\subsection{System Invariants}

To make governance verifiable, DTF enforces four lifecycle invariants. Let $\Omega_t$ denote the execution obligations, and $B_t=(\mu_t,\rho_t,\tau_t,\Omega_t)$ be the maximum authority boundary derived from the proof.

\paragraph{Constraint 1: Proof-bound Execution.}
No governed high-stakes mutation $X_t$ may execute without a corresponding valid proof.
\[
\forall X_t \neq \emptyset, \quad \exists \JP_t \in \mathcal{J} \text{ s.t. } \JP_t = f(I_t, C_t, P_t)
\]

\paragraph{Constraint 2: Consensus-gated Authority.}
No valid Execution Identity may be issued unless the consensus explicitly approves the proof.
\[
\EI_t \neq \emptyset \implies q(A_t, \Gamma_t) = \Approve
\]

\paragraph{Constraint 3: Non-escalation.}
The effective authority scope of $\EI_t$, extracted via $\Scope(\cdot)$, must be contained within the proof-derived boundary.
\[
\forall \EI_t \neq \emptyset, \quad \Scope(\EI_t) \preceq B_t
\]

\paragraph{Constraint 4: Evidence Completeness.}
Every intent proposal, regardless of decision outcome, must map to exactly one structurally complete Evidence Chain.
\[
\forall I_t \in \mathcal{I}, \quad |\EC_t| = 1 \text{ where } \EC_t \text{ captures } (D_t, \EI_t, X_t, O_t)
\]

\subsection{What this paper adds beyond OpenKedge}
\label{sec:openkedge-delta}

OpenKedge contributes one governed mutation substrate: intent-governed mutation, context-aware policy evaluation, execution contracts and task-oriented identities, and lineage through the Intent-to-Execution Evidence Chain (IEEC)~\cite{openkedge2026}. DTF does not replace that substrate. It is a general authorization-verification layer for governed mutation systems, and OpenKedge is the substrate through which we instantiate and evaluate it.

The delta is the object of authorization. OpenKedge governs whether and how an intent may mutate state. DTF defines the stable object that independent evaluators approve ($\JP$), the consensus semantics over that object ($q(A_t,\Gamma_t)$), the proof-derived authority that reaches the execution substrate ($\EI$), and the authorization lifecycle evidence required for later replay ($\EC$). In the OpenKedge instantiation, a task-oriented identity becomes an Execution Identity only when its authority is derived from a Justification Proof and consensus record; an IEEC becomes a DTF Evidence Chain only when it preserves proof, attestations, issuance, execution, and outcome as a replayable authorization lifecycle.

\begin{center}
\centering
\footnotesize
\begin{tabular}{p{0.31\linewidth} p{0.30\linewidth} p{0.30\linewidth}}
\toprule
\textbf{Concern} & \textbf{OpenKedge-based substrate} & \textbf{DTF layer in this paper} \\
\midrule
Mutation unit & Intent-governed mutation proposal & Justification Proof as stable authorization object \\
Policy path & Context-aware policy evaluation & Consensus validation semantics over proof attestations \\
Execution authority & Contracts and task-oriented identities & Proof-derived Execution Identity \\
Lineage & IEEC records intent-to-execution lineage & Evidence Chain records replayable authorization lifecycle \\
\bottomrule
\end{tabular}
\end{center}

\begin{figure}[t]
    \centering
    \begin{tikzpicture}[
        >=Stealth,
        ok/.style={draw=blue!60!black, fill=blue!5, rounded corners=3pt, thick, align=center, text width=1.9cm, minimum height=1.1cm},
        dtf/.style={draw=violet!70!black, fill=violet!6, rounded corners=3pt, thick, align=center, text width=1.9cm, minimum height=1.1cm},
        arr/.style={->, thick, draw=black!70},
        dashedarr/.style={->, dashed, thick, draw=black!50},
        label/.style={ color=black!80, fill=white, fill opacity=0.9, text opacity=1, inner sep=1pt},
        layer/.style={draw, dashed, rounded corners=5pt, thick, inner sep=10pt}
    ]

    \node[ok] (ok_intent) at (0,0) {Intent\\Gateway};
    \node[ok, draw=gray, text=gray, fill=gray!10] (ok_policy) at (4.8,0) {Policy Engine\\(Arbiter)};
    \node[ok, draw=gray, text=gray, fill=gray!10] (ok_contract) at (7.2,0) {Execution\\Contract};
    \node[ok] (ok_exec) at (9.6,0) {Execution\\Substrate};
    \node[ok, draw=gray, text=gray, fill=gray!10] (ok_ieec) at (12.0,0) {IEEC\\(Audit Log)};

    \draw[dashedarr] (ok_intent) -- (ok_policy);
    \draw[dashedarr] (ok_policy) -- (ok_contract);
    \draw[dashedarr] (ok_contract) -- (ok_exec);
    \draw[dashedarr] (ok_exec) -- (ok_ieec);

    \node[dtf] (dtf_jp) at (2.4, 3.2) {Justification\\Proof $\JP_t$};
    \node[dtf] (dtf_swarm) at (4.8, 3.2) {Evaluation\\Swarm $A_t$};
    \node[dtf] (dtf_ei) at (7.2, 3.2) {Execution\\Identity $\EI_t$};
    \node[dtf] (dtf_ec) at (12.0, 3.2) {Evidence\\Chain $\EC_t$};

    \draw[arr] (ok_intent.north) to[out=90, in=180] node[above left=0.05cm, label] {construct} (dtf_jp.west);
    \draw[arr] (dtf_jp.north) to[out=30, in=150] node[above, label] {validate} (dtf_swarm.north);
    \draw[arr] (dtf_swarm.north) to[out=30, in=150] node[above, label] {approve} (dtf_ei.north);
    \draw[arr] (dtf_ei.east) to[out=0, in=90] node[above right=0.05cm, label] {authorize} (ok_exec.north);
    \draw[arr] (ok_exec.north east) -- node[above, label, sloped] {record} (dtf_ec.south west);

    \draw[<->, dotted, thick, draw=violet!80] (ok_policy) -- node[right, label, text=violet!90!black] {maps} (dtf_swarm);
    \draw[<->, dotted, thick, draw=violet!80] (ok_contract) -- node[right, label, text=violet!90!black] {maps} (dtf_ei);
    \draw[<->, dotted, thick, draw=violet!80] (ok_ieec) -- node[right, label, text=violet!90!black] {maps} (dtf_ec);

    \begin{scope}[on background layer]
        \node[layer, draw=blue!40, fill=blue!2, fit=(ok_intent) (ok_policy) (ok_contract) (ok_exec) (ok_ieec), yshift=-5pt] (okbox) {};
        \node[layer, draw=violet!40, fill=violet!2, fit=(dtf_jp) (dtf_swarm) (dtf_ei) (dtf_ec), yshift=5pt] (dtfbox) {};
    \end{scope}

    \node[anchor=north west, font=\bfseries\footnotesize, color=blue!70!black] at ([yshift=-0.1cm]okbox.south west) {Example Substrate (OpenKedge)};
    \node[anchor=south west, font=\bfseries\footnotesize, color=violet!70!black] at ([yshift=0.1cm]dtfbox.north west) {DTF Verification Pipeline};

    \end{tikzpicture}
    \caption{Generic DTF verification pipeline with OpenKedge shown as one substrate mapping. DTF defines proof, evaluator attestations, proof-derived Execution Identity, and Evidence Chain; the OpenKedge row shows the corresponding intent governance, execution contract, task identity, and lineage hooks.}
    \label{fig:dtf-overview}
\end{figure}
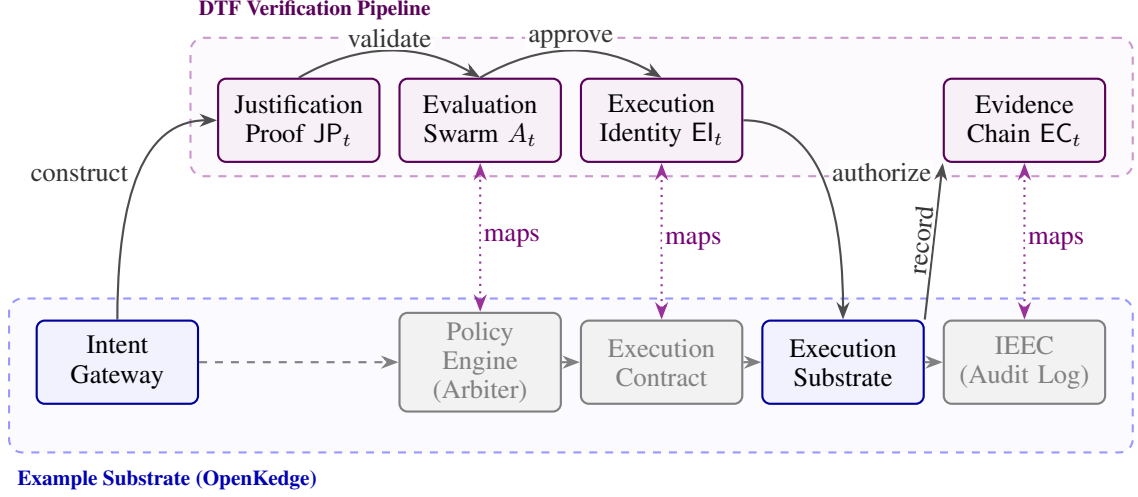

The upper path in Figure~\ref{fig:dtf-overview} is the generic DTF authorization pipeline; the OpenKedge row is one concrete substrate mapping used in this paper.

\section{Justification Proof}

A Justification Proof ($\JP$) is the stable authorization object inspected by evaluators and preserved for replay. It is not a natural-language explanation and not a transcript of the agent's reasoning. It is a structured decision artifact whose contents are sufficient to determine whether a specific mutation can receive bounded authority.

These properties distinguish Justification Proofs from ordinary audit messages or human-readable explanations. Audit messages are typically post-hoc; they record what happened after the fact. Explanations may be persuasive but are often unverifiable. By contrast, a Justification Proof is an authorization artifact whose role is to make the basis and boundary of execution explicit before authority is issued.

\subsection{Structure}

We represent a proof as
\[
\JP_t=(M_t,S_t,\Pi_t,R_t,B_t)\in\mathcal{J},
\]
where $M_t$ is the mutation specification, $S_t$ the bound state and context snapshot, $\Pi_t$ the policy basis, $R_t$ the risk and admissibility assessment, and $B_t$ the maximum execution boundary.

Each field has a narrow role:
\begin{itemize}
    \item $M_t$ states what mutation is proposed.
    \item $S_t$ records the authorization-relevant context used at decision time.
    \item $\Pi_t$ identifies the policies, versions, and constraints applied.
    \item $R_t$ classifies risk and required approval strength.
    \item $B_t=(\mu_t,\rho_t,\tau_t,\Omega_t)$ defines the action ($\mu_t$), resource ($\rho_t$), time ($\tau_t$), and obligation ($\Omega_t$) limits that may be issued if approval succeeds.
\end{itemize}

The proof may reference richer artifacts such as dependency snapshots, simulation results, or model-generated plans, but its authorization-relevant content must be explicit and machine-inspectable.

\subsection{Semantics}

A valid $\JP$ satisfies three properties.

\paragraph{Explicitness.}
Authorization-relevant assumptions must appear in the proof or in referenced immutable artifacts. Hidden model state or informal operator memory is not part of the proof.

\paragraph{Re-evaluability.}
Given the stored proof, evaluators and auditors can revisit the admissibility judgment without reconstructing the proposing agent's internal reasoning.

\paragraph{Boundary derivation.}
The proof determines the maximal authority that could be issued, extracted via the boundary function $\Boundary$:
\[
\Boundary(\JP_t)=B_t.
\]
Approval may narrow this boundary, but issuance may not expand it.

\subsection{Construction}

Proof construction follows four structured steps. Let $\mathsf{Normalize}$, $\mathsf{BindContext}$, $\mathsf{EvaluatePolicy}$, and $\mathsf{DeriveBoundary}$ be state-transition functions governing intent processing. The construction of $\JP_t$ proceeds as follows:
\begin{align*}
I'_t &= \mathsf{Normalize}(I_t) \\
S_t &= \mathsf{BindContext}(I'_t, C_t) \\
(\Pi_t, R_t) &= \mathsf{EvaluatePolicy}(S_t, P_t) \\
B_t &= \mathsf{DeriveBoundary}(\Pi_t, R_t)
\end{align*}
The proof-construction function $f$ from our model is defined as the composition of these steps:
\[
f(I_t, C_t, P_t) \triangleq (M_t, S_t, \Pi_t, R_t, B_t) = \JP_t,
\]
where the normalized intent $I'_t$ serves as the mutation specification $M_t$. In the OpenKedge substrate, this mirrors the intent-governed pipeline but changes the authorization artifact produced by the pipeline. The output is a durable proof object, not a transient ``approve'' or ``deny'' token; independent validators can agree, disagree, or escalate over the same artifact.

\subsection{Example}

For a proposed \texttt{TerminateInstance} mutation, $M_t$ identifies the action and target instance. $S_t$ records relevant topology, traffic, ownership, protection tags, and incident state. $\Pi_t$ identifies policies governing destructive actions and protected resources. $R_t$ may mark the action as critical if the instance is on a dependency path. $B_t$ either permits no issuance or restricts authority to exactly that instance, action, and time window, possibly with an execution-time obligation to re-check traffic.

The proof replaces a vague claim such as ``the instance appears unused'' with a replayable authorization object.

\section{Consensus Validation}

A proof alone is a claim. Consensus validation turns that claim into an authorization decision. DTF does not run full Byzantine consensus over replicated application state; it distributes trust over authorization. The question is whether a proof is sufficient to derive bounded execution authority, and the answer must come from an explicit evaluator set rather than one unchecked component.

One evaluator is insufficient because authorization failures are heterogeneous. A policy evaluator may apply the written rule correctly while missing stale topology. A dependency checker may see current state but lack business-risk context. A simulator may model the ordinary case while missing protected-resource policy. A human reviewer may notice operational intent but miss a malformed proof field. Consensus validation is not decorative voting; it prevents any one failure mode from becoming execution authority.

\subsection{Evaluator Model}

Each evaluator $v_i\in V_t$ receives the same $\JP_t$ and emits an attestation
\[
a_i^t=v_i(\JP_t)\in\mathcal{A},
\]
where $\mathcal{A}$ includes $\Approve$, $\Reject$, and $\Abstain$, plus optional structured objections or obligations. For example, an attestation may be represented as
\[
a_i^t=(v_i,d_i^t,\omega_i^t),
\qquad
d_i^t\in\{\Approve,\Reject,\Abstain\},
\]
where $\omega_i^t$ records evaluator-specific annotations such as objections, obligations, or confidence metadata.

Practical deployments should use heterogeneous evaluator classes:
\begin{itemize}
    \item policy evaluators for explicit governance rules;
    \item state evaluators for topology, dependency, and freshness checks;
    \item risk evaluators for blast radius and reversibility;
    \item simulation evaluators for predicted effects;
    \item human-supervised evaluators for elevated or exceptional cases.
\end{itemize}

Heterogeneity matters only if the consensus rule can observe it. DTF records evaluator class, proof hash, decision, objections, and obligations in $A_t$, and lets $\Gamma_t$ require specific classes, veto rights, quorum thresholds, or escalation on disagreement.

\subsection{Consensus Rule}

The consensus rule encodes quorum thresholds, required evaluator diversity, veto rights, and risk-sensitive escalation. Let $\mathcal{A}^+$ and $\mathcal{A}^-$ denote the sets of approving and rejecting evaluators, respectively:
\begin{align*}
\mathcal{A}^+ &= \{ v_i \in V_t \mid d_i^t = \Approve \} \\
\mathcal{A}^- &= \{ v_i \in V_t \mid d_i^t = \Reject \}
\end{align*}
Let $V_{\text{veto}} \subseteq V_t$ represent the subset of evaluators granted veto authority for the specific mutation class. The governance metadata $\Gamma_t$ supplies the risk-adjusted quorum threshold $k_\Gamma$ and rejection threshold $r_\Gamma$. We define the consensus function $q: \mathcal{A}^n \times \mathcal{G} \rightarrow \mathcal{D}$ as:
\[
q(A_t, \Gamma_t) = 
\begin{cases} 
\Reject & \text{if } \exists v_i \in V_{\text{veto}} : d_i^t = \Reject \text{ or } |\mathcal{A}^-| \ge r_\Gamma \\
\Approve & \text{if } \left(\nexists v_i \in V_{\text{veto}} : d_i^t = \Reject\right) \text{ and } |\mathcal{A}^-| < r_\Gamma \text{ and } |\mathcal{A}^+| \ge k_\Gamma \\
\Escalate & \text{otherwise}
\end{cases}
\]
Higher-risk mutations require stronger quorum limits $k_\Gamma$, greater evaluator diversity within $\mathcal{A}^+$, explicit veto handling, or human review.

Equivalently, issuance readiness is the predicate
\[
\Ready(\JP_t,A_t,\Gamma_t)
\iff
q(A_t,\Gamma_t)=\Approve.
\]

Approval means the proof is authorization-ready under recorded inputs. It does not mean the action is globally optimal. The semantics are deliberately limited: DTF requires authority to come through a checked path; it does not claim complete world knowledge.

\subsection{Failure Handling}

Consensus validation rejects or escalates when quorum is missing, required evaluator classes are absent, the proof becomes stale, evaluator outputs are malformed, the derived boundary is ambiguous, or disagreement itself indicates unsafe uncertainty. Disagreement is not treated as an implementation inconvenience; it is a signal about authorization risk. Emergency handling is modeled as a governed break-glass policy, not a bypass: consensus requirements may change, but proof, narrowed identity, evidence, and post-event review remain mandatory.

\subsection{Audit Value}

Evaluator disagreement is evidence. DTF therefore preserves all attestations, not only the aggregate decision. Later review can distinguish clear approval, contested approval, rejection, escalation, and malformed evaluation. The record also makes the consensus layer accountable: reviewers can see which failure modes were checked, which objections were raised, and whether the configured rule converted that evidence into the appropriate authorization outcome.

\section{Execution Identity}

Execution Identity ($\EI$) is the concrete authority produced by an approved proof. It is the main difference between merely deciding that a mutation is allowed and giving the execution substrate an enforceable, proof-derived scope for that mutation.

Execution Identity changes the locus of authorization from a durable role to a recorded decision. In role-based or attribute-based systems, an agent is assigned an identity (e.g., \texttt{AgentServiceAccount}) and granted broad permissions (e.g., \texttt{TerminateInstance}). When the agent acts, the system checks if the identity has the permission. This assumes the identity is a stable locus of trust.

DTF avoids this assumption. The proposing agent's identity is metadata for the decision, not the source of authority. Trust is placed in the Justification Proof and its validation.

\subsection{Definition}

If $D_t=q(A_t,\Gamma_t)=\Approve$, the system derives
\[
\EI_t=h(\JP_t,A_t,\Gamma_t)\in\mathcal{E}.
\]
An $\EI$ is ephemeral, non-transferable, and scoped to the proof-derived boundary $B_t$. It contains or references:
\begin{itemize}
    \item the permitted action class;
    \item the authorized resource set;
    \item temporal validity bounds;
    \item execution-time obligations;
    \item lineage to $\JP_t$, $A_t$, $\Gamma_t$, and $D_t$.
\end{itemize}

An $\EI$ is not a durable role. It is a per-decision capability whose existence is justified by the recorded proof and approval set.

\subsection{Scope Relation}

Let $\Scope:\mathcal{E}\rightarrow\mathcal{B}$ map an identity to its effective authority boundary. Valid issuance requires
\[
\Scope(\EI_t)\preceq B_t,
\]
where $\preceq$ is the partial order over boundaries: action classes, resource sets, validity intervals, and obligations must be no broader than those encoded in $B_t$. The execution broker may further narrow the scope, for example by reducing lifetime or requiring an additional final precondition check. It may not widen the scope beyond the proof boundary.

\subsection{Enforcement}

The execution substrate must admit a governed mutation $X_t$ only when
\[
\Valid(\EI_t,X_t)=1.
\]
This predicate checks action, resource, time, and obligations. If the target platform cannot enforce a boundary precisely, the broker must add a mediation layer or refuse issuance for that mutation class.

\subsection{Cloud Mapping}

In cloud infrastructure, $\EI$ can be implemented using short-lived credentials such as AWS STS sessions with restrictive session policies and resource conditions~\cite{awsstsdocs}. The credential format is secondary to lineage and containment: the issued authority must be traceable to $\JP_t$ and incapable of authorizing actions outside $B_t$.

\subsection{Contrast with OpenKedge Task-Oriented Identity}

OpenKedge introduces task-oriented identities as the execution-bound mechanism for approved intents~\cite{openkedge2026}. $\EI$ gives that mechanism explicit authorization semantics. A task-oriented identity becomes an Execution Identity when its scope is derived from a Justification Proof, issued only after consensus, and preserved as evidence.

\section{Evidence Chain}

The Evidence Chain is the durable record that makes proof-derived authority replayable. Conventional audit logs can show that an API call occurred. An Evidence Chain must also show why authority was issued, who or what approved it, what boundary was encoded, and whether execution stayed within that boundary.

Logging execution events is insufficient for verifiable infrastructure. If a system logs that an instance was terminated by an agent, an auditor knows \emph{what} happened, but not \emph{why} it was authorized. Did the agent hallucinate the context? Did a policy engine fail? Was the mutation escalated to a human who incorrectly approved it? To answer these questions, the system must record the entire authorization lifecycle. The Evidence Chain records that lifecycle.

\subsection{Record}

For each governed mutation attempt,
\[
\EC_t=(I_t,C_t,P_t,\JP_t,A_t,\Gamma_t,D_t,\EI_t,X_t,O_t).
\]
Implementations may add timestamps, policy versions, hashes, signatures, evaluator identities, state digests, and substrate receipts. These enrich integrity and attribution but do not change the core requirement: authorization and execution must remain in the same lifecycle record.

\subsection{Completeness}

Evidence completeness requires that every governed execution have a record linking proof, approval, derived identity, mutation, and outcome. Rejected and escalated attempts should also be recorded because they explain what the system refused and why. Missing evidence is not an observability gap; it is an authorization failure.

\subsection{Replay}

A complete evidence record should let an auditor answer:
\begin{itemize}
    \item what mutation was proposed;
    \item what context and policy were used;
    \item which evaluators approved, rejected, or abstained;
    \item what boundary was issued;
    \item what execution occurred and what outcome resulted.
\end{itemize}

Replay does not require reconstructing the entire external world. It requires preserving enough authorization-relevant state to re-check the decision under its recorded inputs.

\subsection{Integrity}

The chain should be append-only and tamper-evident. Hash linking, signatures, immutable storage, and correlation with substrate audit systems such as CloudTrail can strengthen the implementation~\cite{awscloudtraildocs}. DTF does not mandate one cryptographic construction; it mandates that proof, approval, identity, and outcome cannot silently drift apart.

\subsection{Relation to IEEC}

OpenKedge's IEEC records the intent-to-execution lineage of governed mutation~\cite{openkedge2026}. DTF refines the evidence requirements for replayable authorization by requiring explicit storage of the Justification Proof, evaluator attestations, governance metadata, and Execution Identity linkage.

\section{Safety and Audit Properties}

This section states system properties of the composed DTF pipeline. The claims below are not properties of any component in isolation, and they should not be read as claims of semantic correctness for every approved action. They hold under the assumptions stated below, when Justification Proofs, consensus validation, Execution Identity, and Evidence Chains are implemented as the ordered authorization path.

These properties are architectural rather than component-local. Traditional systems trust principals and then constrain their actions through policy and logging. DTF makes the decision lifecycle the object of trust: authority emerges from explicit proof, distributed validation, bounded issuance, and durable evidence. This matters in agentic infrastructure, where the initiating reasoning process may be non-deterministic, partially observed, or unreliable.

\subsection{Assumptions}

\paragraph{A1: Enforcement fidelity.}
The execution substrate enforces the action, resource, time, and obligation limits encoded in a valid $\EI$.

\paragraph{A2: Evidence integrity.}
Evidence records cannot be silently modified or deleted after append.

\paragraph{A3: Consensus enforcement.}
The consensus rule approves only when its configured quorum, veto, diversity, and escalation requirements are satisfied.

\paragraph{A4: Proof correspondence.}
The stored proof and context correspond to the mutation attempt being evaluated with the fidelity captured in evidence.

\subsection{System Invariants}

\paragraph{Proof-bound execution.}
Under A1 and A3, any admitted governed mutation $X_t$ has a corresponding $\JP_t$, $A_t$, and $D_t=\Approve$ from which a valid $\EI_t$ was derived. The substrate admits execution only through $\EI$, and $\EI$ is issued only after proof approval.

\paragraph{Consensus-gated authority.}
Under A3, no valid $\EI_t$ is issued unless $D_t=q(A_t,\Gamma_t)=\Approve$. A single evaluator cannot create authority unless the governance rule explicitly defines that as sufficient.

\paragraph{Non-escalation.}
Under A1, if $X_t$ executes through $\EI_t$, then its effective scope is contained within $B_t$. This follows from $\Scope(\EI_t)\preceq B_t$ and substrate enforcement.

\paragraph{Evidence completeness.}
Under A2, each admitted governed mutation remains attached to an evidence record containing intent, context, proof, attestations, governance metadata, identity, execution, and outcome.

\paragraph{Replayable verification.}
Under A2 and A4, the authorization basis of each complete record can be re-examined from stored artifacts. Replayability is about the decision lifecycle, not perfect reconstruction of all external state.

\subsection{Composed Pipeline Property}

Under A1--A4, every admitted governed high-stakes mutation is proof-bound, consensus-gated, scope-bounded, evidence-preserved, and replayable from its lifecycle record.

\paragraph{Proof sketch.}
The DTF pipeline constrains each stage by the previous one:
\[
(I_t,C_t,P_t)\rightarrow \JP_t\rightarrow A_t\rightarrow D_t\rightarrow \EI_t\rightarrow X_t\rightarrow O_t\rightarrow \EC_t.
\]
Approval requires the proof and attestation set; issuance requires approval; execution requires a valid identity; identity scope is contained in the proof boundary; evidence stores the lifecycle. The listed properties follow by composition.

\subsection{Limits}

This composed property does not imply semantic optimality, correct policy design, complete world knowledge, or resilience to total compromise of all evaluators and stores. It establishes a narrower infrastructure property: high-stakes authority cannot be legitimately materialized without recorded proof, recorded approval, bounded issuance, and durable evidence.

\section{Implementation}

DTF is implemented as a verification layer over the OpenKedge-based governed mutation substrate used in our prototype. The prototype binds the abstract functions to this substrate and to AWS primitives, but the decomposition into proof construction, validation, authority derivation, and evidence persistence is substrate-agnostic. Our prototype consists of approximately 4,500 lines of Go and is deployed as a suite of highly available microservices. It interposes on destructive infrastructure operations, privilege changes, and production configuration updates. The implementation materializes the model functions as explicit service boundaries and persisted artifacts rather than as implicit checks inside a single policy engine.

\paragraph{Prototype footprint.}
\begin{center}
\centering
\footnotesize
\begin{tabular}{lrrl}
\toprule
\textbf{Module} & \textbf{LOC} & \textbf{Replicas} & \textbf{Primary model function} \\
\midrule
Intent Gateway & 650 & 3 & $\mathsf{Normalize}$ \\
Justification Engine & 1,200 & 3 & $f$, $\mathsf{BindContext}$, $\mathsf{DeriveBoundary}$ \\
Swarm Coordinator & 700 & 3 & $q(A_t,\Gamma_t)$ \\
Execution Broker & 900 & 3 & $h$, $\Scope$, $\Valid$ \\
Evidence Store Adapter & 500 & 3 & $\EC_t$ persistence \\
Evaluation Harness & 550 & 1 & workload generation and replay \\
\bottomrule
\end{tabular}

\vspace{2pt}
\emph{Replicas are deployed across three availability zones for the online control path.}
\end{center}

\subsection{Function Realization}

The runtime path implements the functions from Section~3 as follows.
\begin{itemize}
    \item $f(I_t,C_t,P_t)$ is implemented by the Intent Gateway and Justification Engine. The gateway canonicalizes $I_t$ into $I'_t$, binds the request to a context snapshot $C_t$, evaluates the applicable policy bundle $P_t$, derives $B_t$, and emits the structured proof $\JP_t=(M_t,S_t,\Pi_t,R_t,B_t)$.
    \item Each evaluator function $v_i(\JP_t)$ is implemented as an isolated worker that consumes the same serialized proof and emits an attributed attestation $a_i^t=(v_i,d_i^t,\omega_i^t)$. Attestations are signed and include evaluator class, input proof hash, decision, objections, and obligations.
    \item The consensus function $q(A_t,\Gamma_t)$ is implemented by the swarm coordinator. It verifies attestation signatures, checks evaluator diversity and freshness, applies veto rules, computes the risk-adjusted quorum from $\Gamma_t$, and returns $\Approve$, $\Reject$, or $\Escalate$.
    \item $\Boundary(\JP_t)$ is materialized as the boundary field $B_t=(\mu_t,\rho_t,\tau_t,\Omega_t)$ embedded in the proof. The broker treats this field as the maximum issuable authority.
    \item $h(\JP_t,A_t,\Gamma_t)$ is implemented by the Execution Broker. It runs only after $q(A_t,\Gamma_t)=\Approve$, converts the approved boundary into a substrate-specific temporary credential, and records lineage to the proof, attestation set, governance metadata, and decision.
    \item $\Scope(\EI_t)$ and $\Valid(\EI_t,X_t)$ are implemented by broker-side admission checks. Before forwarding a mutation, the broker verifies that the effective action, resource, time window, and obligations of the credential are no broader than $B_t$ and that the concrete mutation attempt is authorized by the issued identity.
    \item $\EC_t$ is implemented by the Evidence Store as one append-only lifecycle record containing the intent, bound context digest, policy version, proof, attestations, governance metadata, decision, issued identity reference, attempted mutation, and observed outcome.
\end{itemize}

\subsection{Components}

\paragraph{Intent Gateway and Justification Engine.}
The gateway receives proposed mutations via gRPC and implements $\mathsf{Normalize}$ by resolving aliases, validating schemas, and converting tool-specific calls into canonical mutation specifications. The Justification Engine (approx. 1,200 LOC) implements $\mathsf{BindContext}$ through concurrent state collection from Redis, inventory services, incident state, ownership metadata, and an internal graph database for dependency resolution. It then implements $\mathsf{EvaluatePolicy}$ and $\mathsf{DeriveBoundary}$ by loading a versioned policy bundle, assigning risk class and required governance metadata, and computing the maximum boundary $B_t$. In our deployment, context gathering typically completes in under 20~ms, issuing 9 parallel reads per proof at the median and 17 at the 95th percentile. Serialized proofs have a median size of 6.4~KB and a 95th percentile size of 18.7~KB.

\paragraph{Evaluation Swarm.}
The swarm runs independent evaluators over the same proof and applies $D_t=q(A_t,\Gamma_t)$. Evaluators are implemented as stateless AWS Lambda functions to ensure parallel execution and strict isolation. The deployed evaluator set contains 7 functions across 5 classes: 2 policy evaluators, 2 dependency-state evaluators, 1 risk evaluator, 1 simulation evaluator, and 1 human-escalation adapter. Low-risk mutations require 3 approvals from at least 2 classes; high-risk mutations require 4 approvals from at least 3 classes; protected-resource mutations additionally grant veto authority to the dependency-state and human-escalation classes. Attestations are compact, with a median serialized size of 0.9~KB. The coordinator rejects malformed attestations, escalates missing required evaluator classes, honors veto-capable evaluators for protected mutation classes, and requires risk-sensitive $m$-of-$n$ cryptographic signatures before progressing.

\paragraph{Execution Broker and Evidence Store.}
The Execution Broker derives $\EI_t$ by exchanging the consensus-approved $\JP_t$ for temporary credentials. It translates $B_t$ into substrate-specific action, resource, validity, and obligation constraints, then performs a final $\Scope(\EI_t)\preceq B_t$ check before exposing the credential to the execution path. During execution, the broker evaluates $\Valid(\EI_t,X_t)$ and refuses calls that drift from the approved action, resource, time window, or required obligations. The median identity lifetime is 300~s for destructive actions and 120~s for emergency break-glass actions. The Evidence Store persists $\EC_t$ as an append-only ledger using Amazon QLDB (Quantum Ledger Database) to provide tamper-evident persistence, optionally linking to substrate audit receipts. Each logical lifecycle record is represented by 4 append events on average: proof creation, attestation closure, identity issuance or refusal, and substrate outcome. The median complete evidence record is 11.2~KB.

\subsection{AWS Instantiation}

Our cloud deployment relies heavily on native AWS primitives. The broker implements $h$ by assuming short-lived sessions via AWS Security Token Service (STS), constrained by inline session policies and resource conditions generated from $B_t$~\cite{awsstsdocs}. The generated session policies contain a median of 3 allow statements and 2 explicit resource conditions; no generated policy in the evaluation exceeds 1.8~KB. Temporal scope is enforced through short session duration and broker-side expiry checks. Obligation checks, such as verifying that a target is no longer receiving production traffic, are performed immediately before the substrate call. Evidence records are continuously correlated with CloudTrail logs~\cite{awscloudtraildocs}. In the evaluated workload, this mapping provides containment without requiring custom substrate modifications.

\subsection{Instance Termination Path}

For \texttt{TerminateInstance}(r), the gateway normalizes the request into $M_t=(\texttt{TerminateInstance},r)$. The Justification Engine binds dependency context, traffic state, ownership, protection tags, and incident state into $S_t$; records the governing destructive-action policy in $\Pi_t$; assigns the risk and admissibility assessment $R_t$; and derives $B_t$ as a single-action boundary over resource $r$ with a strict 5-minute validity window and a no-active-traffic obligation. Policy, state, simulation, and risk evaluators attest to this proof in parallel. If $D_t=\Approve$, the broker derives an $\EI_t$ scoped to exactly that termination action on $r$. Before execution, the broker re-checks the obligation and evaluates $\Valid(\EI_t,X_t)$. The Evidence Store records both success and refusal paths with tamper-evident linkage.

\section{Evaluation}

The evaluation is a mechanism validation of DTF in a governed synthetic workload. It asks whether the implementation realizes the model functions---proof construction, independent attestation, consensus, authority derivation, boundary enforcement, execution validity, evidence completeness, and overhead---for high-stakes cloud mutations. Our testbed consists of a simulated environment mimicking a multi-region cloud deployment handling 10,000 synthetic agent mutations over 50 independent runs. Each run contains 200 mutations and replays the same workload against DTF, standing IAM, and policy-only precheck baselines.

Several metrics below are invariant-oriented. A 100\% result for boundary refusal or evidence completeness means that the implementation enforced the designed containment or recording invariant in this controlled workload; it does not imply open-world semantic correctness or universally correct policy. We separate containment metrics, replayability metrics, and policy-quality questions. The evaluation supports claims about proof-derived authority and replayable enforcement under controlled scenarios, not unconstrained internet-scale agent behavior.

\subsection{Questions}

\begin{itemize}
    \item \textbf{EQ1: Proof construction.} Does $f(I_t,C_t,P_t)$ produce a proof containing the mutation, context, policy, risk assessment, and boundary fields needed for independent review?
    \item \textbf{EQ2: Independent validation.} Do evaluator functions $v_i(\JP_t)$ produce attributable attestations, and does $q(A_t,\Gamma_t)$ prevent unilateral unsafe authority derivation?
    \item \textbf{EQ3: Authority containment.} Does $h(\JP_t,A_t,\Gamma_t)$ produce an $\EI_t$ whose effective scope satisfies $\Scope(\EI_t)\preceq B_t$ and reduces authority compared with standing credentials?
    \item \textbf{EQ4: Execution validity and replayability.} Does the broker enforce $\Valid(\EI_t,X_t)$, and can $\EC_t$ reconstruct the authorization lifecycle?
    \item \textbf{EQ5: Overhead.} Is the added latency acceptable for high-risk operations?
\end{itemize}

\subsection{Scenarios}

We evaluate three representative workloads: destructive resource termination (3,500 requests), production configuration mutation during partial degradation (4,500 requests), and emergency break-glass execution (2,000 requests). The workload includes 7,600 admissible requests and 2,400 unsafe or adversarial variants. Unsafe variants include missing dependency context (900), stale state (650), protected-resource targeting (350), post-approval boundary drift (300), and malformed evaluator outputs (200). Baselines are (B1) standing IAM credentials with standard audit logs and (B2) a policy-only precheck that still executes with standing authority. We exclude standard "human-in-the-loop" approval workflows from the latency baselines because human review introduces $O(\text{minutes})$ of latency, which strictly dominates infrastructure overhead and is not directly comparable to automated swarm consensus.

\subsection{Metrics}

The metrics are proof completeness rate, attestation coverage, unsafe mutation block or escalation rate, authority reduction factor, boundary-violation refusal rate, evidence completeness rate, replay success rate, decision latency, and broker overhead. Proof completeness checks whether all fields of $\JP_t=(M_t,S_t,\Pi_t,R_t,B_t)$ are populated and hash-linked to their source artifacts. Attestation coverage checks that required evaluator classes emit well-formed attestations over the same proof hash. Boundary-violation refusal injects attempted action, resource, and time-window drift after approval and measures whether $\Valid(\EI_t,X_t)$ refuses execution.

\paragraph{Aggregate results.}
\begin{center}
\centering
\footnotesize
\begin{tabular}{lrrr}
\toprule
\textbf{Metric} & \textbf{DTF} & \textbf{B1} & \textbf{B2} \\
\midrule
Complete proof records & 100.0\% & 0.0\% & 72.4\% \\
Required attestation coverage & 100.0\% & -- & -- \\
Unsafe block or escalation & 100.0\% & 0.0\% & 86.0\% \\
Boundary-drift refusal & 100.0\% & 0.0\% & 0.0\% \\
Mean mutable resources per approval & 1.0 & 450.0 & 450.0 \\
Evidence completeness & 100.0\% & 41.8\% & 64.6\% \\
Replay success & 99.9\% & 23.0\% & 58.7\% \\
\bottomrule
\end{tabular}
\end{center}

\paragraph{Latency breakdown.}
\begin{center}
\centering
\footnotesize
\begin{tabular}{lrrrr}
\toprule
\textbf{Stage} & \textbf{Mean} & \textbf{p50} & \textbf{p95} & \textbf{p99} \\
\midrule
Proof construction and context binding & 18.5 & 16.8 & 34.7 & 48.9 \\
Parallel evaluator processing & 24.3 & 21.1 & 52.8 & 79.4 \\
Consensus aggregation & 3.1 & 2.7 & 7.6 & 11.8 \\
Broker issuance and STS exchange & 12.4 & 10.9 & 26.5 & 41.2 \\
\midrule
End-to-end decision latency & 58.3 & 52.6 & 112.4 & 171.3 \\
\bottomrule
\end{tabular}
\end{center}

\subsection{Findings}

\paragraph{Safety.}
DTF achieved a 100\% block or escalation rate across all 2,400 unsafe variants in the governed workload. In contrast, B2 (policy-only precheck) allowed 336 unsafe mutations through, corresponding to the 14\% unsafe pass-through rate caused by stale local state or missing dependency context. This result measures enforcement of the configured proof, consensus, and veto rules, not correctness of every possible real-world policy. No single evaluator was able to derive authority without a consensus decision: all 200 malformed evaluator-output cases and all 350 protected-resource veto cases were rejected or escalated by $q(A_t,\Gamma_t)$. In emergency scenarios, DTF successfully processed all 2,000 break-glass requests while constraining the authority to the exact target rather than relying on broad administrative override.

\paragraph{Authority containment.}
$\EI$ reduced the effective execution scope. While standing credentials (B1) exposed an average of 450 resources and a 95th percentile of 1,120 resources to potential mutation per role, DTF constrained authority to exactly 1 approved resource per request, yielding a 99.7\% authority reduction factor at the mean and 99.9\% at the 95th percentile. In 300 boundary-drift tests, attempts to reuse an approved identity for a different action, resource, or expired time window were refused by the broker. For destructive actions, this prevents the tested class of post-approval blast-radius expansion.

\paragraph{Auditability.}
The Evidence Chain achieved a 100\% completeness rate for the DTF lifecycle fields, persisting proof, attestations, governance metadata, decision, issuance reference, attempted mutation, and outcome for all 10,000 attempts. The 10,000 logical records produced 39,812 append events and 10,000 terminal outcome entries. In replay testing, auditors successfully reconstructed the authorization lifecycle for 9,991 events without manual log correlation, compared to 2,300 events under B1 standard audit logs and 5,870 events under B2. Replay failures were attributable to deliberately injected external receipt loss rather than missing DTF lifecycle records.

\paragraph{Overhead.}
The governed path added acceptable latency for high-stakes operations. Proof construction and context gathering averaged 18.5~ms. Evaluator processing via parallel Lambda execution averaged 24.3~ms. Consensus aggregation, signature verification, and quorum evaluation averaged 3.1~ms. Broker execution and STS credential issuance added an overhead of 12.4~ms. The total end-to-end decision latency averaged 58.3~ms, with a median of 52.6~ms, 95th percentile of 112.4~ms, and 99th percentile of 171.3~ms. Even at the 99th percentile, the added delay remains small relative to destructive infrastructure workflows, which are typically dominated by substrate operation latency and human review time.

\subsection{Ablations}

Removing consensus leaves a proof object but reintroduces unilateral evaluator failure: in the malformed-output workload, single-evaluator approval would have allowed 61 of 200 malformed cases to proceed. This matches the swarm design rationale: heterogeneous evaluator disagreement matters only if it can block or escalate authority derivation. Removing $\EI$ preserves decision records but loses execution containment: all 300 post-approval boundary-drift attempts become executable because the substrate still sees broad standing authority. Removing the Evidence Chain preserves runtime gating but reduces replay success from 99.9\% to 52.1\% because auditors must reconstruct proof, approval, and outcome from separate logs. The mechanisms are complementary rather than interchangeable.

\subsection{Limitations}

The prototype evaluation tests high-risk mutations rather than benchmarking every possible agent action. Its containment and replayability results reflect architecture-controlled scenarios with known mutation classes, explicit policies, and injected failure modes; they should not be read as unconstrained internet-scale benchmarks. Results depend on the fidelity of captured context, the correctness of policy, the quality and independence of evaluator classes, and the granularity with which the substrate can enforce $\EI$ boundaries. We do not claim resilience to total compromise of all evaluators, broker, and evidence storage. The evidence supports the narrower claim that proof-derived authority is practical and improves containment and replayability for governed infrastructure mutations.

\section{Discussion and Future Work}

DTF changes the object of authorization from a durable principal to a recorded decision lifecycle. That shift matters because agentic systems are non-deterministic and partially observed: the system need not trust the agent's internal reasoning if execution authority can be derived only from explicit proof, independent approval, and bounded issuance.

\subsection{Tradeoffs}

DTF adds latency, storage, and operational complexity. These costs are justified for high-stakes mutations but should not be imposed indiscriminately on routine reads or low-risk operations. Evidence fidelity also creates a storage tradeoff: richer context improves replay, while compact records reduce operational burden. Evaluator diversity improves resilience but increases governance design work.

\subsection{Limitations}

DTF does not synthesize correct policy, guarantee globally optimal decisions, or solve prompt injection inside the proposing agent. Its value depends on context fidelity: if dependency state, ownership metadata, incident state, or policy versions are stale or incomplete, the proof may faithfully encode the wrong basis for authorization. It also depends on evaluator diversity quality. A nominal swarm whose evaluators share the same data source, model failure mode, or operational blind spot provides less protection than its quorum size suggests.

DTF also assumes that the governed path is actually interposed, that $\EI$ boundaries are enforceable at the substrate granularity required by the mutation, and that evidence integrity is maintained. Some substrates expose coarse permissions that may require a broker or mediation layer before DTF can enforce narrow scopes. DTF is not designed to survive total compromise of the evaluator set, broker, and evidence store. These are infrastructure and governance assumptions, not model capabilities.

\subsection{Future Work}

Future work includes typed proof languages for $\JP$, adaptive evaluator selection based on risk and historical disagreement, stronger cross-domain Evidence Chains for workflows spanning organizations, simulation-backed approval for irreversible operations, and structured human review for high-assurance escalation.

\section{Conclusion}

OpenKedge motivates and instantiates the governed mutation substrate used in this paper: intent, context, policy, bounded execution, and lineage. The paper's contribution is DTF, a verification model for making authorization proof-bound, consensus-gated, scope-bounded, and replayable across governed agentic infrastructure. It introduces Justification Proofs, consensus validation, Execution Identity, and Evidence Chains as the mechanisms that turn an approved mutation into replayable authority.

The invariant can be stated compactly: no high-stakes execution without proof, no authority without consensus, and no valid mutation detached from evidence. DTF makes that invariant an executable authorization lifecycle for agentic infrastructure, rather than an audit aspiration after the fact.

\bibliographystyle{unsrtnat}
\bibliography{references}

\clearpage
\appendix
\section{Notation}
\begin{table}[h!]
\centering
\footnotesize
\begin{tabular}{p{0.34\linewidth} p{0.58\linewidth}}
\toprule
\textbf{Symbol} & \textbf{Meaning} \\
\midrule
$\mathcal{I},\mathcal{C},\mathcal{P}$ & Intent, context, and policy spaces. \\
$\mathcal{J},\mathcal{A},\mathcal{G}$ & Proof, attestation, and governance-metadata spaces. \\
$\mathcal{B},\mathcal{E}$ & Execution-boundary and Execution Identity spaces. \\
$\mathcal{X},\mathcal{O}$ & Mutation-attempt and execution-outcome spaces. \\
$I_t\in\mathcal{I}$ & Canonical mutation intent at logical time $t$. \\
$C_t\in\mathcal{C}$ & Bound context snapshot used for authorization. \\
$P_t\in\mathcal{P}$ & Applicable policy bundle and version. \\
$f:\mathcal{I}\times\mathcal{C}\times\mathcal{P}\to\mathcal{J}$ & Proof-construction function. \\
$\JP_t\in\mathcal{J}$ & Justification Proof derived from $(I_t,C_t,P_t)$. \\
$V_t=\{v_1,\ldots,v_n\}$ & Evaluator set selected for the mutation at time $t$. \\
$a_i^t\in\mathcal{A}$ & Attestation emitted by evaluator $v_i$ over $\JP_t$. \\
$A_t=(a_1^t,\ldots,a_n^t)$ & Ordered attestation record. \\
$\Gamma_t\in\mathcal{G}$ & Governance metadata, including quorum, veto, and escalation rules. \\
$q:\mathcal{A}^{n}\times\mathcal{G}\to\mathcal{D}$ & Consensus function. \\
$D_t\in\mathcal{D}$ & Consensus decision, where $\mathcal{D}=\{\Approve,\Reject,\Escalate\}$. \\
$B_t=(\mu_t,\rho_t,\tau_t,\Omega_t)$ & Proof-derived execution boundary: action, resources, time, obligations. \\
$\Boundary:\mathcal{J}\to\mathcal{B}$ & Map from a proof to its maximal execution boundary. \\
$h:\mathcal{J}\times\mathcal{A}^{n}\times\mathcal{G}\to\mathcal{E}$ & Authority-derivation function. \\
$\EI_t\in\mathcal{E}$ & Execution Identity derived from an approved proof. \\
$\preceq$ & Componentwise containment order over execution boundaries. \\
$\Scope:\mathcal{E}\to\mathcal{B}$ & Map from an identity to its effective authority boundary. \\
$\Scope(\EI_t)\preceq B_t$ & Issued identity scope is no broader than the approved boundary. \\
$\Valid:\mathcal{E}\times\mathcal{X}\to\{0,1\}$ & Predicate deciding whether an identity authorizes a mutation attempt. \\
$X_t\in\mathcal{X}$ & Attempted or executed mutation. \\
$O_t\in\mathcal{O}$ & Observed execution outcome. \\
$\EC_t$ & Evidence record linking proof, approval, authority, execution, and outcome. \\
\bottomrule
\end{tabular}
\caption{Core notation. Authority is modeled as a computed state derived from proof and consensus, not as a standing property of a caller.}
\label{tab:notation}
\end{table}

\end{document}